\documentclass[sigconf]{acmart}
\usepackage{multirow}
\usepackage[ruled,vlined]{algorithm2e}
\usepackage{subfigure}
\usepackage{amsmath}
\usepackage{url}
\usepackage{enumitem}
\usepackage[american]{babel}

\renewcommand\footnotetextcopyrightpermission[1]{} % removes footnote with conference information in first column
\AtBeginDocument{%
  \providecommand\BibTeX{{%
    \normalfont B\kern-0.5em{\scshape i\kern-0.25em b}\kern-0.8em\TeX}}}

\setcopyright{acmcopyright}
\copyrightyear{2020}
\acmYear{2020}
\acmConference[preprint '20]{preprint '20}{August 03--07, 2020}{San Diego, CA}
\acmBooktitle{preprint '20: ACM SIGKDD Conference on Knowledge Discovery and Data, August 03--07, 2020, San Diego, CA}
\acmPrice{15.00}
\acmISBN{978-1-4503-XXXX-X/18/06}

\begin{document}

\title{Learning from a Lightweight Teacher for Efficient\\ Knowledge Distillation}

\author{Yuang Liu}
\affiliation{%
  \institution{East China Normal University}
  \streetaddress{3663 North Zhongshan Rd}
  \city{Shanghai}
  \country{China}}
\email{frankliu624@gmail.com}

\author{Wei Zhang}
\authornote{Corresponding Author.}
\orcid{1234-5678-9012}
\affiliation{%
  \institution{East China Normal University}
  \streetaddress{3663 North Zhongshan Rd}
  \city{Shanghai}
  \country{China}}
\email{zhangwei.thu2011@gmail.com}

\author{Jun Wang}
\affiliation{%
  \institution{East China Normal University}
  \streetaddress{3663 North Zhongshan Rd}
  \city{Shanghai}
  \country{China}}
\email{wongjun@gmail.com}

\renewcommand{\shortauthors}{Liu and Zhang, et al.}

\begin{abstract}
Knowledge Distillation (KD) is an effective framework for compressing deep learning models, realized by a student-teacher paradigm requiring small student networks to mimic the soft target generated by well-trained teachers.
However, the teachers are commonly assumed to be complex and need to be trained on the same datasets as students.
This leads to a time-consuming training process.
The recent study shows vanilla KD plays a similar role as label smoothing and develops teacher-free KD, being efficient and mitigating the issue of learning from heavy teachers.
But because teacher-free KD relies on manually-crafted output distributions kept the same for all data instances belonging to the same class, its flexibility and performance are relatively limited.
To address the above issues, this paper proposes en efficient knowledge distillation learning framework LW-KD, short for lightweight knowledge distillation. 
It firstly trains a lightweight teacher network on a synthesized simple dataset, with an adjustable class number equal to that of a target dataset.
The teacher then generates soft target whereby an enhanced KD loss could guide student learning, which is a combination of KD loss and adversarial loss for making student output indistinguishable from the output of the teacher.
Experiments on several public datasets with different modalities demonstrate LW-KD is effective and efficient, showing the rationality of its main design principles.
\end{abstract}





\maketitle

\section{Introduction}
\begin{figure}[!t]
  \centering
  \subfigure[KD]{\label{subfig:KD-sketch}
  \includegraphics[width=.47\linewidth]{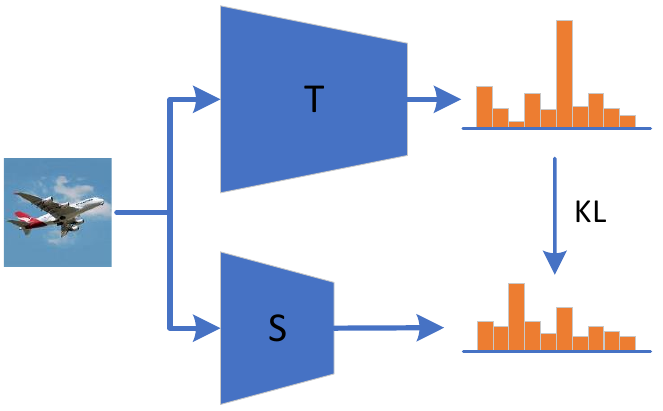}
  %\caption{fig1}
  }
  %\quad
  \subfigure[KD-shuffle]{\label{subfig:KD-S-sketch}
  \includegraphics[width=.47\linewidth]{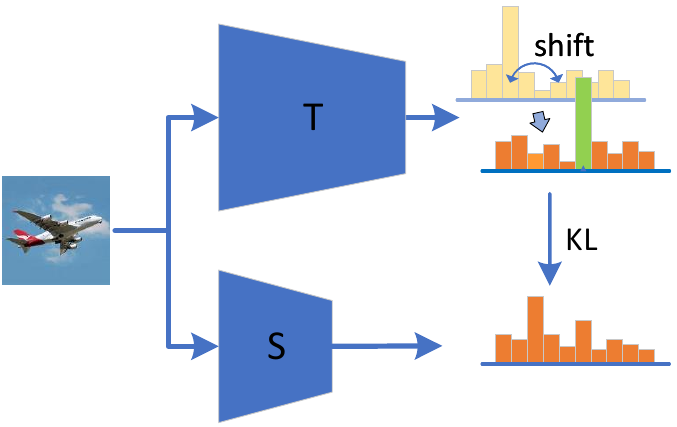}
  }
  \caption{The sketches of KD and its shuffled variant. T corresponds to teacher and S for student. We use the trapezium area to denote the teacher is larger than the student.}
  \label{fig:kd-framework}
\end{figure}

Modern deep learning models gain tremendous success due to the design of complicated multi-layer neural networks (e.g., ResNet~\cite{he2016deep}), the collection of large-scale datasets, and the training with more effective optimization techniques (e.g., Adam~\cite{KingmaB14}) and computational-intensive resources (e.g., GPU).
However, this paradigm is not so applicable in the era of Mobile Internet and Internet Of Things (IOT).
This is because low-end intelligent terminal devices dominant the market, with small storage capacity and low computing power.
As a result, there is an urgent need to train portable neural networks while maintaining the effectiveness of big models as much as possible. 
Knowledge Distillation (KD), firstly proposed by Hinton et al.~\cite{hinton2015distilling}, is an elegant learning framework to address this need.
This is realized by taking soft target generated by one or more powerful but complicated teacher networks as another target (beyond ground-truth) to train compact student model, as shown in Figure~\ref{subfig:KD-sketch}.
So far, KD has been exploited in computer vision (e.g., image classification~\cite{romero2014fitnets} and semantic segmentation~\cite{LiuCLQLW19}), natural language processing (e.g., machine translation~\cite{ChenLCL17} and relation classification~\cite{VyasC19}), and recommender system~\cite{TangW18}, to name a few. 

However, the common assumption of KD, leveraging strong teacher networks, still suffers from a non-negligible computational burden, due to the training of the complex teacher models in the training process.
The costly process is inevitable for each specific situation or dataset, 
This issue is even amplified with the trends of proposing larger models for better performance, such as VGG19~\cite{karen15very} and BERT~\cite{devlin2019bert}. 
As reported in the study~\cite{devlin2019bert}, the big Transformer model should be trained for 3.5 days on 8 NVIDIA P100 GPU, not to mention the industrial applications.
Hence it brings large energy consumption and causes environmental loss~\cite{StrubellGM19}.

Recently, teacher-free knowledge distillation (Tf-KD) is developed~\cite{yuan2019revisit} to free students from learning complicated teachers.
The authors have demonstrated that vanilla KD which utilizes soft output class distributions of teachers plays a role of label smoothing ~\cite{szegedy2016rethinking,muller2019does} to constrain the class predictions of students when training them.
To further verify this, we conduct a verification experiment on the CIFAR datasets (cf. Section~\ref{subsec:data}) by removing the ``dark knowledge'' contained in the soft targets.
Specifically, as shown in Figure~\ref{subfig:KD-S-sketch}, KD-shuffle is designed to perform shuffle operation on each instance in a training set, whereby all the elements of the teacher output, except the one with maximal probability, are randomly permuted.
Table~\ref{tab:intro} shows that in most cases, shuffling the soft targets do not cause the performance of the student network declines notably. 
As such, the conclusion that vanilla knowledge distillation is similar to label smoothing~\cite{yuan2019revisit} is reasonable.

To achieve efficient knowledge, Tf-KD adopts a virtual teacher model with manually-crafted output class distributions as transferred knowledge, eliminating the necessity of training a real and complex teacher.
In particular, they assign a pre-defined larger value to the ground-truth class of an instance and make other classes share the same small probability value.
Thus the crafted distributions are class-dependent and share some similar spirit with label smoothing which uses a global uniform class distribution.
However, although Tf-KD is efficient, the manual distributions are kept the same for all data examples belonging to the same class.
This causes that the flexibility of Tf-KD is relatively limited.
As shown in the experimental section, it is hard for Tf-KD
to reach the overall performance level of vanilla KD.
This phenomenon naturally poses a major challenge: how to achieve efficient knowledge distillation inspired by Tf-KD while maintaining classification efficacy like vanilla KD.

\begin{table}[!t]
  \caption{Results of KD and its shuffled variant (student: ResNet18~\cite{he2016deep}).}
  \label{tab:intro}
  \begin{tabular}{cccc}
    \toprule
    Dataset & Teacher & KD & KD-shuffle\\
    \midrule
    \multirow{3}*{CIFAR10} & ResNet20~\cite{he2016deep} & 92.98 & 92.85\\
    ~ & MobileNetV2~\cite{sandler2018mobilenetv2} & 91.20 & 91.33\\
    ~ & ShuffleNetV2~\cite{ma2018shufflenet} & 92.17 & 91.15\\
    \hline
    \multirow{3}*{CIFAR100} & ResNet20~\cite{he2016deep}  & 68.74 & 68.91\\
    ~ & MobileNetV2~\cite{sandler2018mobilenetv2} & 69.00 & 69.15\\
    ~ & ShuffleNetV2~\cite{ma2018shufflenet} & 71.43 & 71.58\\
  \bottomrule
\end{tabular}
\end{table}

In this paper, to address the challenge, we devise a novel efficient knowledge distillation learning framework LW-KD, short for lightweight knowledge distillation.
In this framework, in contrast to vanilla KD, a general lightweight network is leveraged as the teacher, which is even much smaller than student networks to be used.
Since the teacher network has lower capacity than students, it is intuitive for LW-KD to train the teacher on a simple dataset for ensuring satisfied performance, instead of training on the same target dataset used by the student,
To be specific, LW-KD automatically builds a synthetic dataset SynMNIST based on the simple digit dataset MNIST\footnote{\url{http://yann.lecun.com/exdb/mnist/}}, with an adjustable class number equal to that of the target dataset.
It is worth noting that the target dataset could be with a different modality compared with the image modality of SynMNIST.
Based on this, LW-KD trains the teacher to reach a suitable state to generate soft output class distributions.
The framework performs a slight modification to the distribution so as to obtain transferred knowledge.
An enhanced KD loss is developed, involving standard KD loss and adversarial loss, to exploit the knowledge for student training and make the output distributions of the student indistinguishable from the teacher's.

Our key contributions are summarized as the following: 
\begin{itemize}
    \item 
    We indicate the limitations of both vanilla KD which suffers from heavy computational burden in the training process and teacher-free KD which lacks of flexibility.
    To achieve an effective trade-off, we propose to realize KD by learning from a lightweight teacher.
    
    \item
    We propose the novel KD framework named LW-KD.
    It enables training a lightweight teacher on the synthesized simple dataset SynMNIST which could satisfy different requirements of class number.
    We further devise the enhanced KD loss for LW-KD to leverage the output class distributions of the teacher to guide the student learning.
    
    \item 
    We conduct extensive experiments on different types of data, involving image, text, and video.
    The results demonstrate the benefits of LW-KD over vanilla KD and Tf-KD, with better performance than teacher-free KD and sometimes even outperforming vanilla KD, and much faster than vanilla KD in teacher training.
    As a byproduct, the rationality of the main principles in LW-KD is validated. 
    
\end{itemize}

\section{Related Work}

Since the aim of this paper is to learn from a lightweight teacher for knowledge distillation to obtain a well-performed student, we review the literature from the following aspects.

% 1. Knowledge Distillation
\vspace{0.2em} \noindent \textbf{Knowledge Distillation.} 
Transferring knowledge from a large network to a small network is a long-standing topic and has drawn much attention in recent years.
The pioneering work~\cite{hinton2015distilling} has shown that distillation works well for transferring knowledge from an ensemble or a strong and complicated model into a small and compact model. 
The main claim is that using soft targets as complementary to hard targets could carry some useful information among classes learned by teachers.
Due to the efficacy of KD in retaining classification performance and model compression, an enormous amount of research efforts have been spawned.

The first category includes the methods those simply transfer knowledge contained in soft output by teacher models to student models as vanilla KD~\cite{hinton2015distilling}. 
For instance, Lopez-Paz et al.~\cite{Lopez-PazBSV15} unified distillation and privileged information into one framework.
In ~\cite{li2017learning}, Li et al. developed a new framework to learn from noisy labels, by leveraging the knowledge gained from a small clean dataset and semantic knowledge graph to correct the noisy labels. 
You et al.~\cite{YouX0T17} averaged the soft targets generated by multiple teachers as richer knowledge for more effective student learning.
Tan et al.~\cite{TanRHQZL19} associated one teacher network with each source-target language pair for machine translation. 
There are all also some self-KD methods~\cite{xie2019self,zhang2019your,liu2020regularizing} using soft targets generated by a student itself.
But the training process becomes more complicated since the coupling of optimizations on both the teacher and student sides. 

A second line of studies are attributed to the category that transfers structural knowledge obtained from a teacher model.
RKD~\cite{park2019relational} uses distance-wise and angle-wise distillation losses those penalize the structural differences in relations.
Similar with RKD, IRG~\cite{liu2019knowledge} transformation is proposed to model the feature space transformation across layers.
It models three kinds of knowledge, including instance features, instance relationships, and feature space transformation.
SP~\cite{tung2019similarity} exploits a pairwise similarity preserving constraint in distillation loss, computed on each mini-batch.
Both VID~\cite{ahn2019variational} and CRD~\cite{Tian-ICLR20} consider maximizing mutual information between the two networks as a knowledge distillation task. 
Mutual information is leveraged to maximize the variational lower bound in VID and the contrastive loss in CRD, respectively. 

A third category of approaches learns the knowledge revealed in the intermediate layers of teachers.
NST~\cite{huang2017like} treats it as a distribution matching problem and matches the distributions of neuron selectivity patterns between teacher and student networks. 
Romero et al.~\cite{romero2014fitnets} distilled from a teacher using additional linear projection layers to train a relatively narrower student. 
Instead of mimicking a teacher's output activations, Zagoruyko et al.~\cite{komodakis2017paying} proposed to force a student to mimic a teacher's attention maps.
Crowley et al.~\cite{crowley2018moonshine} compressed a model by grouping convolution channels of the model and training it with an attention transfer. 
In~\cite{yim2017gift}, the flow of solution procedure (FSP), generated by computing the Gram matrix of features across layers, is employed for knowledge transfer. 

Although much progress has been made for KD, a very recent study shows that vanilla KD still behaves well compared with other representative KD methods~\cite{Tian-ICLR20}.
To sum up, most of the above KD approaches require one or more large and high-performance teachers.
As aforementioned, the training procedure of teachers is costly for the consideration of time and computational resources.
Although the teacher-free KD framework frees the student learning from relying on powerful teachers, its manually-crafted class distributions are relatively limited, making room for improving student classification performance.
This paper addresses both the limitations of vanilla KD and teacher-free KD by proposing to learn from a lightweight teacher for efficient knowledge distillation.

% 1. Label Smoothing
\vspace{0.2em} \noindent \textbf{Label Smoothing.}
The label smoothing mechanism is firstly proposed in ~\cite{szegedy2016rethinking} to regularize the classifier layer by estimating the marginalized effect of label dropout
during training. 
In fact, it encourages the model to be less confident and makes it more generalizable. 
Label smoothing has been successfully utilized to improve the accuracy of deep learning models across a range of tasks, including image classification~\cite{szegedy2016rethinking}, speech recognition~\cite{chorowski2016towards}, and machine translation~\cite{vaswani2017attention}.
Recently, M{\"u}ller et al.~\cite{muller2019does} summarized and explained several observations when training deep neural networks with label smoothing. 
Yuan et al.~\cite{yuan2019revisit} demonstrated the relation between label smoothing and knowledge distillation, indicating their similar roles in effect.

% 3. Model Compression
\vspace{0.2em} \noindent \textbf{Model Compression and Acceleration.}
Knowledge distillation could be regarded as one branch of model compression methods through transfer learning~\cite{crowley2018moonshine,bhardwaj2019efficient,shu2019co}.
The aim of model compression and acceleration is to create networks with fast computation speed and small parameter complexity.
Meanwhile, they should maintain high performance. 
A straightforward way to achieve this is to design a powerful but lightweight network since the original convolution network has many redundant parameters.
MobileNet~\cite{sandler2018mobilenetv2} is designed with depth-wise separable convolution to replace standard convolution. 
In ShuffleNet~\cite{ma2018shufflenet}, point-wise group convolution and channel shuffle are proposed to reduce the burden of computation while maintaining high accuracy. 
Another manner is network pruning which boosts the speed of inference by pruning the neurons or filters with low importance based on certain criteria~\cite{li2016pruning,zhou2019accelerate}. 
Besides, some other studies exploit low rank approximation to large layers~\cite{sainath2013low} and quantization seeks to use low-precision model parameter representation~\cite{wu2016quantized}.

\section{Methodologies}
In this section, we first revisit the key parts of knowledge distillation, followed by the elaboration of our proposed framework LW-KD.

\subsection{Knowledge Distillation}
The general idea of knowledge distillation is to let a student network to mimic the soft target generated by a teacher, as shown in Figure~\ref{subfig:KD-sketch}.
This is realized in the KD learning framework by adding another loss function to complement the standard cross-entropy loss measuring the gap between ground-truth and predictions.
The added loss constrains the soft output of the student to be similar with the soft output of the teacher, which could be regarded as knowledge learned by the teacher.
Specifically, the loss is defined as follows:
\begin{equation}
  \label{eq:kd-kl}
  \mathcal{L}_{KL}=\frac{1}{|\mathcal{X}^{tr}|}\sum_{i=1}^{|\mathcal{X}^{tr}|} KL(p^s_{\tau}, p^t_{\tau})\,,
\end{equation}
where $KL(\cdot)$ denotes the KL-divergence measure.
$|\mathcal{X}^{tr}|$ is assumed to be the number of instances in the training set $\mathcal{X}^{tr}$.
$p^s_{\tau}$ and $p^t_{\tau}$ are the soft output of the student and soft target generated by the teacher, respectively.

The soft outputs are computed based on the logits, gotten from last layers of neural networks before feeding to the $Softmax$ function.
Taking $p^t_{\tau}$ as an example, it is computed by:
\begin{equation}
  \label{eq:kd-soft-target}
  p_{\tau}^{t}(k)=\frac{\exp \left(z_{k}^{t} / \tau\right)}{\sum_{i=1}^{K} \exp \left(z_{i}^{t} / \tau\right)}\,,
\end{equation}
where $z_{k}^{t}$ is the logit from the teacher and $K$ is the total number of classes.
$\tau$ is a hyperparameter (referred as temperature in ~\cite{hinton2015distilling}) to control the scale of logits.
For the teacher-free KD framework, the major difference is that $p_{\tau}^{t}(k)$ is not calculated based on a well-trained teacher, but is obtained through manually-crafted class distributions.

% \begin{equation}
%   \mathcal{L}_{KD}=(1-\alpha) \mathcal{H}(q, p)+\alpha \mathcal{L}_{K L}\left(p_{\tau}^{t}, p_{\tau}\right)
% \end{equation}

\begin{figure}[!t]
  \centering
  \includegraphics[width=\linewidth]{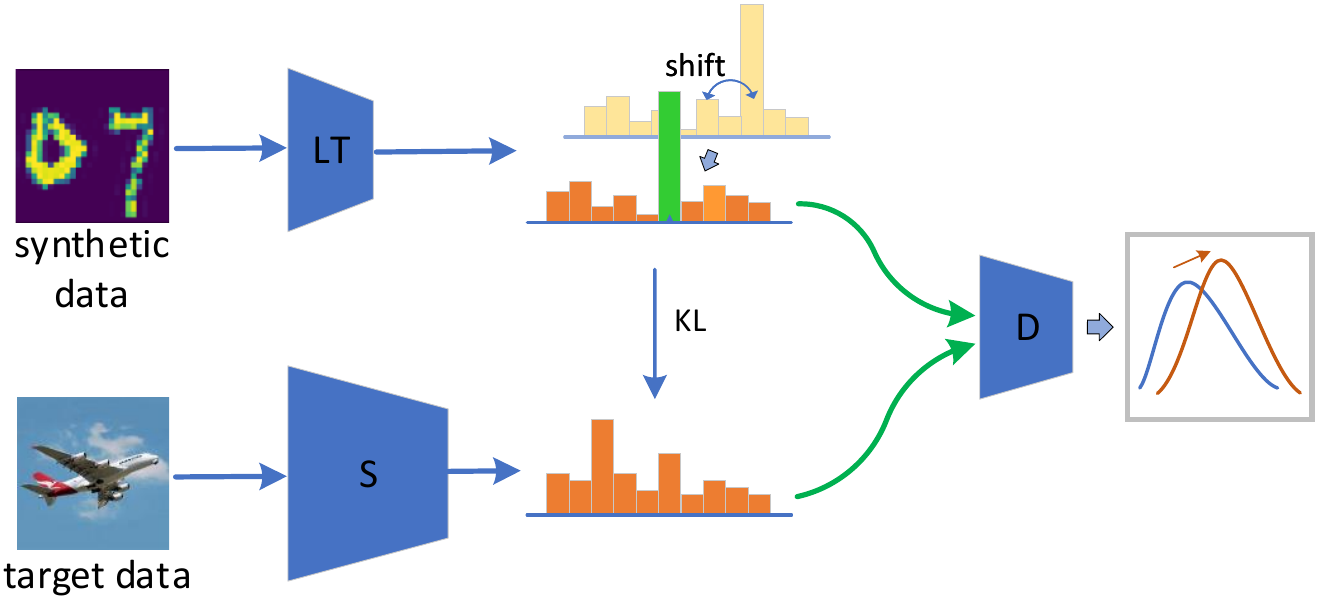}
  \caption{The sketch of the LW-KD learning framework. LT corresponds to a lightweight teacher.}
  \label{fig:LW-KD-framework}
\end{figure}

\subsection{The LW-KD Learning Framework}

\subsubsection{Overview}
Figure~\ref{fig:LW-KD-framework} depicts the concrete procedures of the LW-KD learning framework.
For a target dataset with a specific number of classes, LW-KD firstly synthesizes a tailored dataset based on a simple image dataset.
Afterwards, a lightweight teacher is trained on the synthetic dataset.
Later, the soft output class distributions generated by the teacher are slightly modified to be used in the enhanced KD loss to train a student network on the target dataset.
Under this way, the student could reach its better performance.

In what follows, we mathematically specify the details of LW-KD. 
We use $\mathcal{T}(x;\theta^t)$ to represent the teacher network with trainable parameters $\theta^t$ and $\mathcal{S}(x;\theta^s)$ to denote the student network with learnable parameters $\theta^s$.

\subsubsection{Data Synthesis}
We automatically synthesize the dataset for training a lightweight teacher based on the simple dataset MNIST.
MNIST is composed of fixed-sized images about size-normalized handwritten digits.
The task of classifying the digit images is relatively easy compared with other image classification tasks like CIFAR.
The pioneering study~\cite{lecun1998gradient} indicates the simple neural network LeNet5 could already achieve very low error rate.
As such, it is reasonable for a lightweight teacher to train on this dataset and gain satisfied performance.
%Note that the learning framework LW-KD is generalizable and could use other simple datasets for data synthesis as well.

\begin{algorithm}[!t]
  \caption{The algorithm for synthesizing the SynMNIST dataset.}
  \label{alg:synmnist}
  \SetAlgoLined
  \LinesNumbered
  \KwData{Official MNIST dataset $(\mathcal{X}^{mn}, \mathcal{C}^{mn})$.}
  \KwIn{Number of classes $K$ and number of instances $E_k$ in eacher class $k\in\{1,2,\ldots,K\}$.}
  \KwOut{SynMNIST dataset ($\mathcal{X}^{syn}, \mathcal{C}^{syn}$) for an $N$-class classification task.}

  \For{$n=0$ to $K-1$}{
    Generate a digit label list $L$ by splitting the number $k$\;
    \For{$e=1$ to $E_k$}{
      Get a group of images $I$ from $\mathcal{X}^{mn}$ by referring to the digits in $L$ and classes in $\mathcal{C}^{mn}$\;
      $\hat{I} \gets$ concat($I$)\;
      $x^{syn} \gets$ resize($\hat{I}$)\;
      Append label $k$ to $\mathcal{C}^{syn}$\;
      Append data $x^{syn}$ to $\mathcal{X}^{syn}$\;
    }
  }
\end{algorithm}

% examples of promnist
\begin{figure}[!t]
  \centering
  \subfigure[03]{
  \includegraphics[width=0.8cm]{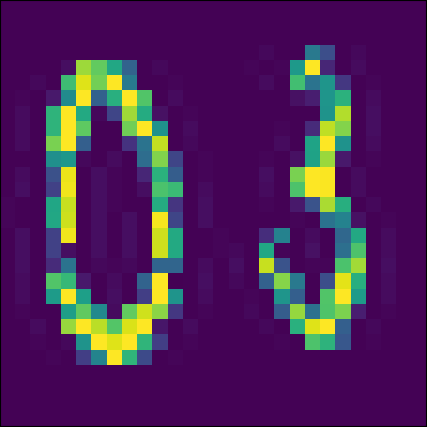}
  %\caption{fig1}
  }
  \quad
  \subfigure[07]{
  \includegraphics[width=0.8cm]{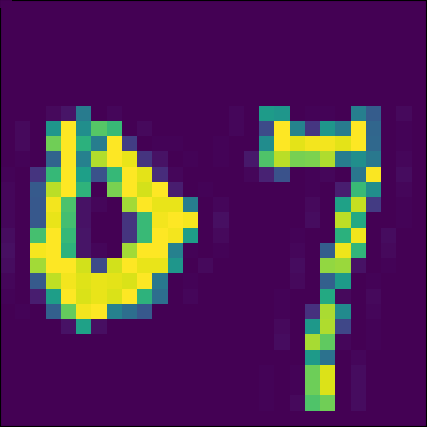}
  }
  % \quad
  % \subfigure[19]{
  % \includegraphics[width=0.8cm]{promnist/19.png}
  % }
  \quad
  \subfigure[30]{
  \includegraphics[width=0.8cm]{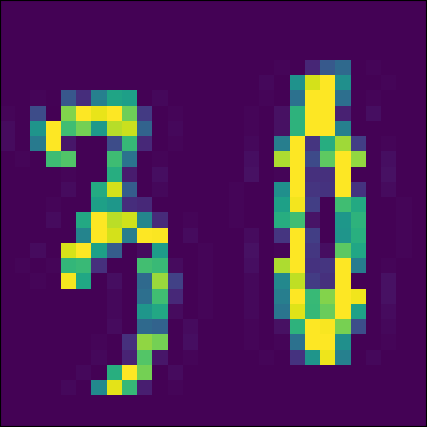}
  }
  \quad
  \subfigure[41]{
  \includegraphics[width=0.8cm]{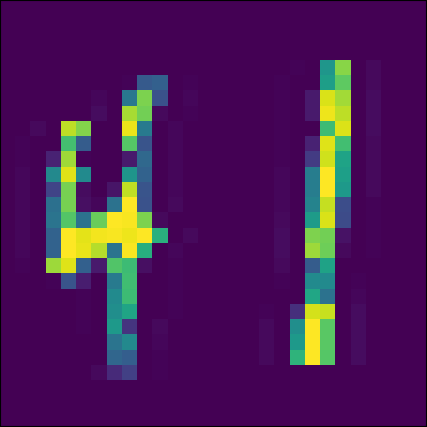}
  }
  \quad
  \subfigure[57]{
  \includegraphics[width=0.8cm]{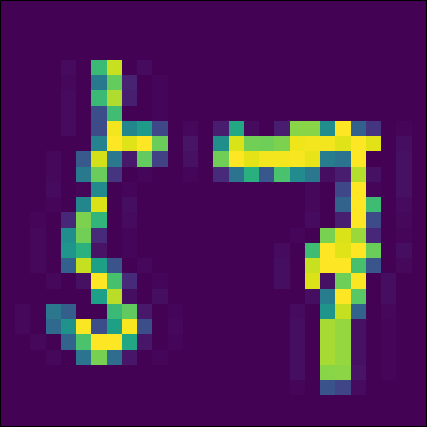}
  }
  \quad
  \subfigure[99]{
  \includegraphics[width=0.8cm]{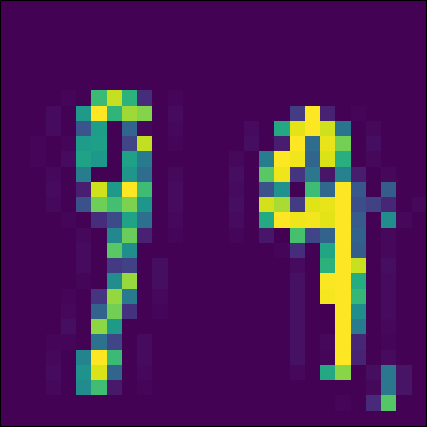}
  }
  \quad

  \subfigure[010]{
  \includegraphics[width=0.8cm]{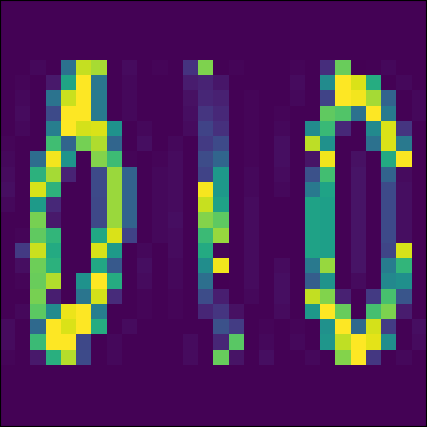}
  }
  \quad
  \subfigure[057]{
  \includegraphics[width=0.8cm]{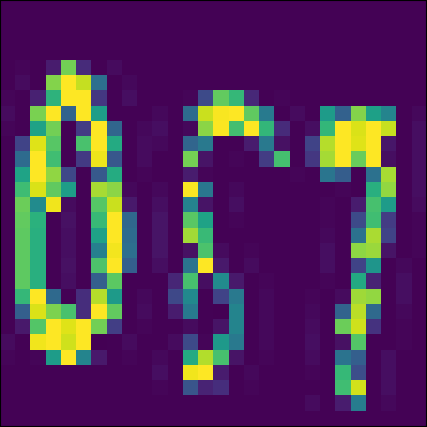}
  }
  \quad
  \subfigure[130]{
  \includegraphics[width=0.8cm]{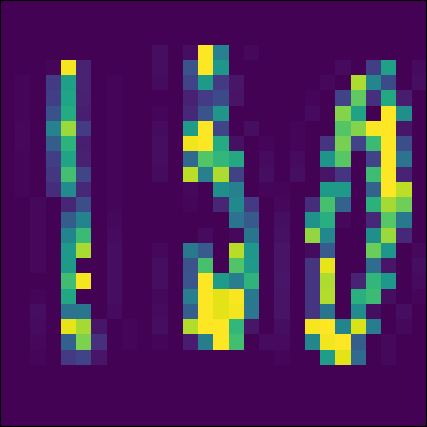}
  }
  % \quad
  % \subfigure[161]{
  % \includegraphics[width=0.8cm]{promnist/161.png}
  % }
  \quad
  \subfigure[169]{
  \includegraphics[width=0.8cm]{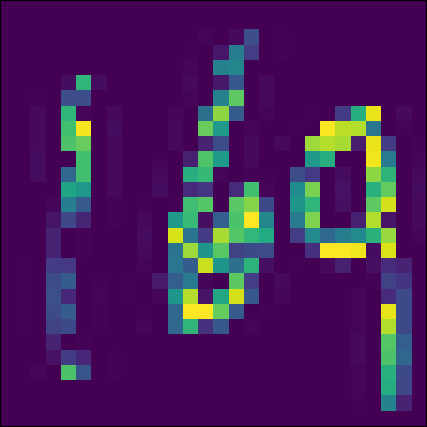}
  }
  \quad
  \subfigure[374]{
  \includegraphics[width=0.8cm]{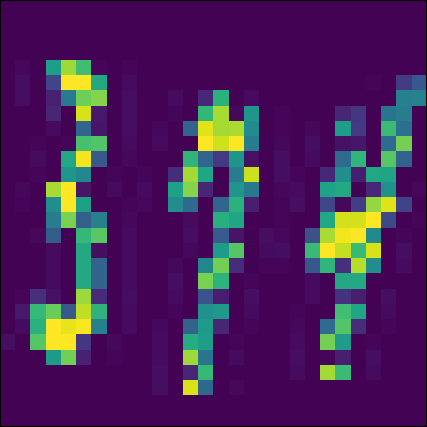}
  }
  \quad
  \subfigure[748]{
  \includegraphics[width=0.8cm]{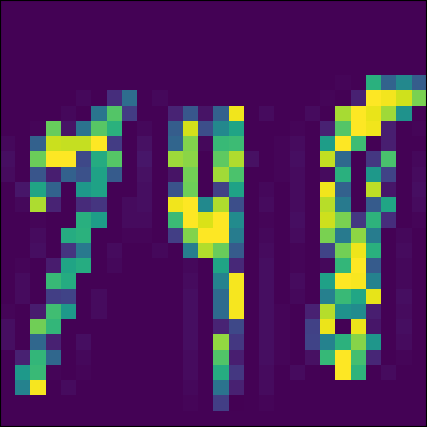}
  }
  \caption{Examples in the SynMNIST dataset.}
  \label{fig:synmnist}
\end{figure}

Algorithm~\ref{alg:synmnist} illustrates the main procedures of constructing the SynMNIST dataset.
Regarding a specific target dataset to be used for training the student $\mathcal{S}$, the algorithm only needs to know its total class number $K$ and the number of instances for each class.
The innovation of the algorithm lies into combining different basic digit images to synthesize new images corresponding to larger numerical values, each of which could denote a specific class.
As a result, this algorithm supports different number of classes.
Figure~\ref{fig:synmnist} shows the synthesized images by Algorithm~\ref{alg:synmnist} for 100-category and 1000-category classification tasks, respectively.
For example, the image ``03'' represents an instance in  the 100-category (``00''-``99'') dataset, corresponding to the fourth class.

\subsubsection{Soft Target Generation}
Given the synthetic dataset ($\mathcal{X}^{syn}, \mathcal{C}^{syn}$), LW-KD trains the lightweight teacher network $\mathcal{T}$ to a good and stable state.
Now the teacher network could generate the soft target for the synthetic dataset: $p^{lt}_{\tau}(k|x^{syn})$ ($k \in \{1...K\}$), w.r.t. the synthetic data instance $x^{syn}$.
The main aim of LW-KD is to transfer the knowledge contained in the probability distributions to benefit the student learning.
Nevertheless, one fundamental issue is that the dataset used for the teacher and the student are dramatically different.
Therefore their classes could not be semantically aligned.
In such a situation, a preliminary thought is that it is impossible to realize effective knowledge transfer.
However, as we emphasized in this paper, vanilla KD acts like label smoothing.
And the aim of LW-KD is to utilize the flexible class distributions generated by real teachers for label smoothing.
Thus LW-KD does not need the strict semantic alignment between classes of the two datasets.

The only modification for the generated soft targets $p^{lt}(k|x^{syn})$ ($k \in \{1...K\}$) is to associate the largest probability value within them with the ground-truth class of a target data instance $x^{tr}$, while leaving other probabilities unchanged.
To this end, we define the following way to update the soft targets:
\begin{equation}
  \label{eq:shift}
  \hat{p}^{lt}_{\tau}(k|x^{tr})=\left\{\begin{array}{ll}
  {p^{lt}_{\tau}(k|x^{syn})} & {\text { if } k \neq c \text{ or } m,} \\
  {\text{shift}(p^{lt}_{\tau}(c|x^{syn})\,, p^{lt}_{\tau}(m|x^{syn}))} & {\text { otherwise}\,,}
  \end{array}\right.
\end{equation}
where $c$ is the real class of the target instance.
$m$ is obtained by $m=\mathop{\arg\max}_{k}(p^{lt}_{\tau}(k|x^{syn}))$.
The operation of shift($\cdot,\cdot$) means swapping the values of the two positions. 
Through the above manner, the soft target generated by teacher becomes somewhat realistic for the target example.

\subsubsection{Enhanced KD Loss for Optimization}
After obtaining the soft target from the lightweight teacher $\mathcal{T}$, we could adopt the standard KD loss adopted by vanilla KD.
It is composed of a cross-entropy loss and a KL-divergence loss (shown in Equation~\ref{eq:kd-kl}), which is defined as follows:
\begin{equation}\label{eq:kd-loss}
  \mathcal{L}_{KD}=(1-\alpha) \mathcal{H}(p, p^s)+\alpha \mathcal{L}_{K L}(\hat{p}_{\tau}^{lt}, p^s_{\tau})\,,
\end{equation}
where $p$ is the ground-truth class distribution and $\alpha$ is a hyperparameter to balance the two losses.
The above equation enables the instance-level guidance from one soft target distribution for training the student on a target instance.
With the consideration of vanilla knowledge distillation as label smoothing, LW-KD goes further by introducing corpus-level guidance, i.e., making the soft class distributions generated by the student indistinguishable from those of the teacher
This is realized by the effective generative adversarial networks (GANs)~\cite{goodfellow2014generative}.

GANs have been widely applied for generating samples satisfying the distribution of real data.
A specific GAN consists of a generator $\mathcal{G}$ to generate desired data and a discriminator $\mathcal{D}$ to identify differences between real instances and generated ones.
To be specific, given an input noise vector $z$, $\mathcal{G}$ maps $z$ to the desired data $x$, i.e., $\mathcal{G}: z \to x$. On the other hand, $\mathcal{D}$ outputs a probability denoting an example to be real, i.e., $\mathcal{D}: x \to [0,1]$.
The objective function of a standard GAN is formulated as below:
\begin{equation}\label{eq:GAN}
\begin{aligned} 
  \mathcal{L}_{GAN}(z, y)=& 
  \mathbb{E}_{y \sim p_{\text{data}(y)}}[\log \mathcal{D}(y)] \\ 
  &+\mathrm{E}_{z \sim p_{z}(z)}[\log (1-\mathcal{D}(\mathcal{G}(z))]\,,
\end{aligned}
\end{equation}
where the generator will be adjusted according to the training error produced by $\mathcal{D}$ using the back propagation strategy. And the optimal generator is:
\begin{equation}
  G^{*}=\arg \min _{\mathcal{G}} \max _{\mathcal{D}} \mathcal{L}_{GAN} \,.
\end{equation}

Here we regard the soft class distributions generated by the teacher network as the real data examples and correspond the student network to the generator $\mathcal{G}$.
A two-layered fully-connected neural network is adopted as the discriminator $\mathcal{D}$.
Upon this, we define the follow objective in our situation:
\begin{equation}
  \label{eq:L_adv}
  \mathcal{L}_{ADV}= \mathbb{E}_{\hat{p}_{\tau}^{lt}\sim\mathcal{T}}[\log \mathcal{D}(\hat{p}_{\tau}^{lt})] +\mathbb{E}_{p_{\tau}^{s}\sim\mathcal{S}}\log [(1-\mathcal{D}(p_{\tau}^s))]\,.
\end{equation}
Through adversarial training, the student can learn from the lightweight teacher better, because the above objective encourages the student to be more confident in the soft target distributions which act as a good regularization.

\begin{algorithm}[!t]
  \caption{The learning algorithm of LW-KD.}
  \label{alg:training}
  \SetAlgoLined
  \LinesNumbered
  \KwData{Synthetic dataset SynMNIST $(\mathcal{X}^{syn}, \mathcal{C}^{syn})$ and the specific training data $(\mathcal{X}^{tr}, \mathcal{C}^{tr})$}
  \KwIn{Lightweight teacher model $\mathcal{T}(x;\theta^t)$ and discriminator $\mathcal{D}(z;\theta^d)$}
  \KwOut{Student model $\mathcal{S}(x;\theta^s)$}
  Quickly pretrain the lightweight teacher on SynMNIST and get a satisfied teacher model $\mathcal{T}(x;\theta^t)$\; 
  Randomly initialize the student model $\mathcal{S}(x;\theta^s)$\;
  Build a simple one-to-one mapping between real instances and synthetic examples\;
  \For{number of training epochs}{
    \For{number of training steps}{
      Select a training instance $(x^{tr}, c^{tr}) \sim (\mathcal{X}^{tr}, \mathcal{C}^{tr})$\;
      Load the mapped instance from synthetic data: $x^{syn} \sim \mathcal{X}^{syn}$\;
      $p^{lt}(c|x^{syn}) \gets \mathcal{T}(x^{syn}; \theta^t)$\;
      $p^s(c|x^{tr}) \gets \mathcal{S}(x^{tr}; \theta^s)$\;
      Compute $p^{lt}_{\tau}$ and $p^s_{\tau}$ in a similar fashion as Eq.~\ref{eq:kd-soft-target}\;
      Perform shift operation (Eq.~\ref{eq:shift}) to get $\hat{p}^{lt}_{\tau}(k|x^{tr})$\;
      Update $\theta^s$ by minimizing Eq.~\ref{eq:Loss}\;
      Update $\theta^d$ by maximizing Eq.~\ref{eq:L_adv}\;
    }
  }
\end{algorithm}

In the end, the enhanced KD loss for LW-KD is formulated as:
\begin{equation}
  \label{eq:Loss}
  \begin{aligned}
    \mathcal{L}&= \mathcal{L}_{KD} + \beta \mathcal{L}_{ADV}\,,
  \end{aligned}
\end{equation}
where $\beta$ is a hyperparameter for balancing the KD loss and the adversarial loss. 
Algorithm~\ref{alg:training} summarizes the overall learning procedure of LW-KD. 
As shown in Line 6 to 13, the knowledge learned by the teacher network on the synthetic dataset is transferred to guide the student learning on the target dataset.

%% ProMNIST
\begin{table*}[!t]
  \caption{Teacher instances of LeNet5 and its variant LeNetW. We use SynMNIST-NUM to denote the class number of the synthesized dataset to be NUM. Each teacher instance is trained on the associated dataset.}
  \label{tab:letnet5-teacher}
  \begin{tabular}{cc|c|c|cccc}
    \toprule
    Type & Name & Model & \#Params & Dataset & \#Example & \#Epoch & Acc \\
    \midrule
    \multirow{4}*{Lightweight teacher} & $\mathcal{T}_1$ & \multirow{4}*{LeNet5~\cite{lecun1998gradient}} & \multirow{4}*{$\sim$61.7K} & MNIST & 60000 & 10 & 98.80 \\
    ~& $\mathcal{T}_2$ & ~ & ~ & SynMNIST-15 & 75000 & 10 & 98.46 \\
    ~& $\mathcal{T}_3$ & ~ & ~ & SynMNIST-100 & 60000 & 10 & 97.22 \\
    ~& $\mathcal{T}_4$ & ~ & ~ & SynMNIST-101 & 101000 & 2 & 84.81\\
    \hline
    \multirow{2}*{Variant for multi-channel image} & $\mathcal{T}_5$ & \multirow{2}*{LeNetW} & \multirow{2}*{$\sim$25.8M} & CIFAR10 & 60000 & 20 & 74.52 \\
    ~& $\mathcal{T}_6$ & ~ & ~ & CIFAR100 & 60000 & 20 & 40.95 \\ 
    \bottomrule
  \end{tabular}
\end{table*}

% Stu on C10
\begin{table*}[!t]
  \caption{Results of image classification for different learning methods. $\diamond$ denotes for the two teachers of each student, we use the more complicated one to compute the average results for its better performance. * represents LW-KD is significantly better than Tf-KD under t-test ($p\leq 0.05$). }
  \label{tab:s_c10}
  \begin{tabular}{c|ccc|cc|cccc}
    \toprule
    Dataset & Student & \#Params & Sup-Stu & Teacher & KD & LSR & Tf-KD & KD[$\mathcal{T}_5$] & Ours[$\mathcal{T}_1$] \\
    \midrule
    \multirow{8}*{CIFAR10} & \multirow{2}*{ResNet20~\cite{he2016deep}} & \multirow{2}*{$\sim$ 69.7K} & \multirow{2}*{92.29} & ResNet18 & 92.98 & \multirow{2}*{92.37} & \multirow{2}*{92.55} & \multirow{2}*{92.46} & \multirow{2}*{92.94} \\
    ~ &~ & ~ & ~ & ResNet50 & \textbf{93.04} & ~ & ~ & ~ \\
    \cline{2-10}
    ~ & \multirow{2}*{MobileNetV2~\cite{sandler2018mobilenetv2}} & \multirow{2}*{$\sim$2.3M} & \multirow{2}*{89.74} & ResNet18 & 91.20 & \multirow{2}*{89.85} & \multirow{2}*{89.96} & \multirow{2}*{90.04} & \multirow{2}*{91.57} \\
    ~ & ~ & ~ & ~ & ResNet50 & \textbf{91.69} & ~ & ~ & ~ \\ \cline{2-10}
    ~ & \multirow{2}*{ShuffleNetV2~\cite{ma2018shufflenet}} & \multirow{2}*{$\sim$1.3M} & \multirow{2}*{91.21} & ResNet18 & 92.17 & \multirow{2}*{91.28} & \multirow{2}*{91.39} & \multirow{2}*{91.51} & \multirow{2}*{92.48} \\
    ~ & ~ & ~ & ~ & ResNet50 & \textbf{92.55} & ~ & ~ & ~ \\ 
    \cline{2-10}
    ~ & VGG19~\cite{karen15very} & $\sim$39.0M & 93.51 & DenseNet121 & 93.75 & 93.58 & 93.30 & -- & \textbf{93.81} \\
    ~ & GoogLeNet~\cite{szegedy2015going} & $\sim$6.3M & 95.17 & ResNeXt29 & 95.21 & 94.73 & 94.62 & -- & \textbf{95.35} \\
    \midrule
    \multirow{8}*{CIFAR100} &\multirow{2}*{ResNet20~\cite{he2016deep}} & \multirow{2}*{$\sim$275.5K} & \multirow{2}*{68.38} & ResNet18 & 68.74 & \multirow{2}*{68.74} & \multirow{2}*{69.04} & \multirow{2}*{68.73} & \multirow{2}*{\textbf{69.39}} \\
    ~ & ~ & ~ & ~ & ResNet50 & 68.78 & ~ & ~ & ~ \\ 
    \cline{2-10}
    ~ & \multirow{2}*{MobileNetV2~\cite{sandler2018mobilenetv2}} & \multirow{2}*{$\sim$2.4M} & \multirow{2}*{67.44} & ResNet18 & 69.00 & \multirow{2}*{67.78} & \multirow{2}*{68.21} & \multirow{2}*{67.77} & \multirow{2}*{\textbf{69.13}} \\
    ~ & ~ & ~ & ~ & ResNet50 & 68.04 & ~ & ~ & ~ \\ 
    \cline{2-10}
    ~ & \multirow{2}*{ShuffleNetV2~\cite{ma2018shufflenet}} & \multirow{2}*{$\sim$1.4M} & \multirow{2}*{70.83} & ResNet18 & 71.43 & \multirow{2}*{70.87} & \multirow{2}*{71.46} & \multirow{2}*{71.24} & \multirow{2}*{\textbf{71.57}} \\
    ~ & ~ & ~ & ~ & ResNet50 & 71.11 & ~ & ~ & ~ \\ 
    \cline{2-10}
    ~ & VGG19~\cite{karen15very} & $\sim$39.3M & 73.40 & DenseNet121 & 73.91 & 73.50 & 73.87 & -- & \textbf{74.18} \\
    ~ & GoogLeNet~\cite{szegedy2015going} & $\sim$6.3M & 79.37 & ResNeXt29 & \textbf{79.58} & 78.83 & 78.16 & -- & 79.44 \\ \midrule
    AVG & -- & -- & 82.13 & -- & $82.77^{\diamond}$ & 82.15 & 82.26 & -- & \textbf{82.99*} \\
    \bottomrule
  \end{tabular}
\end{table*}

\section{Experimental Setup}
This section clarifies the detailed setup for the experiments, including the used datasets, the adopted baseline learning approaches, and the implementations of the proposed LW-KD framework.

\subsection{Datasets}\label{subsec:data}

To ensure reliable comparison, we adopt multiple datasets, covering the modalities of image, text, and video. 

\noindent \textbf{CIFAR10} and \textbf{CIFAR100}. 
CIFAR is a popular image classification benchmark, involving 32$\times$32 RGB images and being widely used in for testing KD.
Both CIFAR10 and CIFAR100 share the same data source which contains 50k training images and 10K testing images.
The difference of the two datasets is that CIFAR10 has 10 classes while CIFAR100 has 100 classes.

% \noindent \textbf{ImageNet.} The 1,000-class dataset from ILSVRC 2012~\cite{ILSVRC15} provides 1.2 million images for training, and 50,000 for validation.

\noindent \textbf{THUCNews\footnote{\url{http://thuctc.thunlp.org/}}}.
THUCNews is a Chinese text classification dataset collected from Sina\footnote{\url{https://www.sina.com.cn/}} during the period from 2005 to 2011.
We extracted 200,000 news headlines from THUCNews with 10 categories, including Finance, Real Estate, Stocks, Education, Technology, Society, Current Affairs, Sports, Games, and Entertainment.
Each category has 20,000 headlines and the text length is between 20 and 30.
For ease of training, validation, and testing, this dataset is segmented by the ratio of 18 to 1 to 1.

\noindent \textbf{ToutiaoNews\footnote{\url{https://github.com/skdjfla/toutiao-text-classfication-dataset}}}.
This dataset was collected from Toutiao\footnote{\url{https://www.toutiao.com/}}, a streaming news platform.
It has 382,688 text instances belonging to 15 categories, e.g. Story, Culture, and Entertainment.
The same segmentation procedure for THUCNews is applied for ToutiaoNews.

\noindent \textbf{UCF101.} UCF101~\cite{Soomro2012ucf101} is currently the largest dataset about human action. 
It consists of 101 action classes, over 13k clips, and 27 hours of video data collected from Youtube.

\subsection{Baselines}
We adopt the following learning approaches directly relevant to LW-KD for comparison.

\begin{itemize}[leftmargin=*]

\item \textbf{Sup-Stu:} 
This is the general name for a group of student models, each of which is trained on a target dataset by supervised learning, but without the help of knowledge distillation learning.

\item \textbf{LSR:} 
It denotes that a student is trained with label smoothing.
Following~\cite{szegedy2016rethinking,muller2019does}, we adopt a global uniform class distribution to smooth the ground-truth distribution.

\item \textbf{Tf-KD:} 
The teacher-free learning framework~\cite{yuan2019revisit} designs a virtual perfect teacher for soft target generation.
The generated distributions by the teacher is class-dependent, sharing a similar spirit with the smoothing distribution used in LSR.

\item \textbf{KD:} 
This represents vanilla KD~\cite{hinton2015distilling} proposed to leverage the knowledge contained in the teachers' output logits.
Although this method is simple compared with its follow-up approaches, its performance is still very competitive compared to other representative KD approaches~\cite{Tian-ICLR20}.

\end{itemize}

For fair comparison, all the adopted approaches are tuned on the validation datasets for achieving their own good performance.

\subsection{Implementation Details}
Our learning framework is implemented with PyTorch~\cite{paszke2019pytorch} on two NVIDIA GTX 2080Ti GPUs. For the part of teacher training, we use SGD with the learning rate of 0.1, momentum of 0.9, and weight decay of 5e-4. 
% And the same setting is applied for the student training.
For adversarial training of the discriminator, we use Adam with learning rate 0.1.
The batch size is set to 64 for UCF101, and 128 for the other datasets.
The lightweight teachers adopted by LW-KD are presented in Table~\ref{tab:letnet5-teacher}.
Othe detailed settings, including the hyperparameters of  $\alpha$, $\beta$, and $\tau$, and the choices of teachers for vanilla KD, are reported for specific datasets.

\section{Experimental Results}
In this part, we aim to address the following research questions:

\begin{itemize}
    \item[\textbf{\texttt{RQ1}:}] How well does LW-KD perform on different datasets compared with the existing vanilla KD and teacher-free KD learning methods?
    
    \item[\textbf{\texttt{RQ2}:}] Do the main design principles involved in LW-KD ensure model effectiveness and efficiency? 
\end{itemize}

We adopt the metric of accuracy (Acc for short) to quantify the classification performance.
Furthermore, we utilize floating point operations (FLOPs), denoting the total calculation cost for one instance, and model size for efficiency analysis.

% -------------------------------------
%% Text -- THUCNews, ToutiaoNews
%--------------------------------------

% Text Table
\begin{table*}[!t]
  \caption{Results of text classification for different learning methods. }
  \label{tab:Text-Result}
  \begin{tabular}{c|c|c|ccccc}
    \toprule
    Student & \#Params & Dataset & Sup-Stu & LSR & Tf-KD & KD[BERT] & Ours\\
    \midrule
    \multirow{2}*{TextCNN~\cite{kim2014convolutional}} & \multirow{2}*{$\sim$2.1M} & THUCNews & 90.58 & 91.33 & 91.38 & 91.78 & \textbf{91.82} [$\mathcal{T}_1$]\\
    ~ & ~ & ToutiaoNews & 85.93 & 86.23 & 86.04 & 86.32 & \textbf{86.65} [$\mathcal{T}_2$]\\
    \hline
    \multirow{2}*{TextRCNN~\cite{lai2015recurrent}} & \multirow{2}*{$\sim$2.6M} & THUCNews & 91.11 & 91.51 & 91.31 & 91.60 & \textbf{91.86} [$\mathcal{T}_1$] \\
    ~ & ~ & ToutiaoNews & 84.89 & 85.36 & 85.02 & 85.80 & \textbf{85.86} [$\mathcal{T}_2$]\\
    \hline
    \multirow{2}*{DPCNN~\cite{johnso2017deep}} & \multirow{2}*{$\sim$1.8M} & THUCNews & 91.07 & 91.29 & 90.66 & 91.51 & \textbf{91.60} [$\mathcal{T}_1$]\\
    ~ & ~ & ToutiaoNews & 84.45 & 84.56 & 84.52 & \textbf{85.35} & 85.02 [$\mathcal{T}_2$]\\
    \hline
    \multirow{2}*{FastText~\cite{joulin2017bag}} & \multirow{2}*{$\sim$152.0M} & THUCNews & 92.32 & 92.66 & 92.33 & 92.47 & \textbf{92.78} [$\mathcal{T}_1$]\\
    ~ & ~ & ToutiaoNews & 86.51 & 87.21 & 86.79 & \textbf{87.62} & 87.46 [$\mathcal{T}_2$]\\
    \hline
    \multirow{2}*{BERT~\cite{devlin2019bert}} & \multirow{2}*{$\sim$102.3M} & THUCNews & 94.59 & 94.63 & 94.37 & 94.77 & \textbf{94.83} [$\mathcal{T}_1$]\\
    ~ & ~ & ToutiaoNews & 89.29 & 89.42 & 89.36 & 89.64 & \textbf{89.77} [$\mathcal{T}_2$]\\
    \bottomrule
  \end{tabular}  
\end{table*}

\subsection{Approach Performance Comparison (\textbf{\texttt{RQ1}})}

\subsubsection{Performance on Image Modality}
We first test the performance on the image classification datasets, i.e., CIFAR10 and CIFAR100.
$\mathcal{T}_1$ and $\mathcal{T}_3$ in Table~\ref{tab:letnet5-teacher} are taken as the two lightweight teachers exploited by LW-KD for CIFAR10 and CIFAR100, respectively.
Since the original form of LeNet5 does not support to model multi-channel images in CIFAR, we design the wider variants of LeNet5, i.e., $\mathcal{T}_5$ and $\mathcal{T}_6$ for later analysis.
Besides, vanilla KD adopts some other complex teachers with size and accuracy described in Table~\ref{tab:complex_teacher_cifar}.
In LW-KD, we set $\alpha=0.5, \beta=0.1$, and $\tau=6.0$ for CIFAR10 and $\alpha=0.1, \beta=0.1$ and $\tau=20.0$ for CIFAR100 through simple grid search.
The learning rate of SGD is decayed by 0.1 at the epochs of 100 and 150, respectively.
The total number of training epochs is set to 200 for the two CIFAR datasets.

% ----------------------------------------
%% Image -- CIFAR10 & 100, ImageNet
% ----------------------------------------
% Teachers for CIFAR
\begin{table}[!t]
  \caption{Strong teacher models trained on CIFAR.}
  \label{tab:complex_teacher_cifar}
  \begin{tabular}{cccc}
    \toprule
    Teacher & \#Params & CIFAR10 & CIFAR100 \\
    \midrule
    ResNet18~\cite{he2016deep} & $\sim$11.2M & 95.18 & 78.18 \\
    ResNet50~\cite{he2016deep} & $\sim$23.5M & 95.53 & 78.86 \\
    DenseNet121~\cite{huang2017densely} & $\sim$7.0M & 95.67 & 79.86 \\
    ResNeXt29~\cite{xie2017aggregated} & $\sim$89.6M & 95.74 & 81.02 \\
    \bottomrule
  \end{tabular}
\end{table}

Table~\ref{tab:s_c10} shows the results on CIFAR10 and CIFAR100, from which we have the following key findings:
\begin{itemize}[leftmargin=*]
\item Tf-KD and LSR gain near performance on the image datasets, which confirms to the expectation since their intrinsic mechanisms are similar.
Compared to KD, although the two learning methods achieve comparable performance on CIFAR100, their results on CIFAR10 are inferior.
This demonstrates that the performance of Tf-KD and LSR might be limited by their inflexible manually-crafted class distributions.

\item Our learning framework LW-KD shows comparable performance with vanilla KD, and outperforms Tf-KD and LSR significantly in most cases.
This observation is meaningful and welcomed since the adopted lightweight teachers $\mathcal{T}_1$ and $\mathcal{T}_3$ are much smaller (about 61.7K in size) than the complicated teachers and are even smaller than the student models.
Besides, using the variant of lightweight teacher trained on the CIFAR datasets does not perform as well as LW-KD.

\end{itemize}

\begin{figure*}[!t]
  \centering
  \subfigure[The student ResNet20 on CIFAR100.]{
  \includegraphics[width=0.4\textwidth]{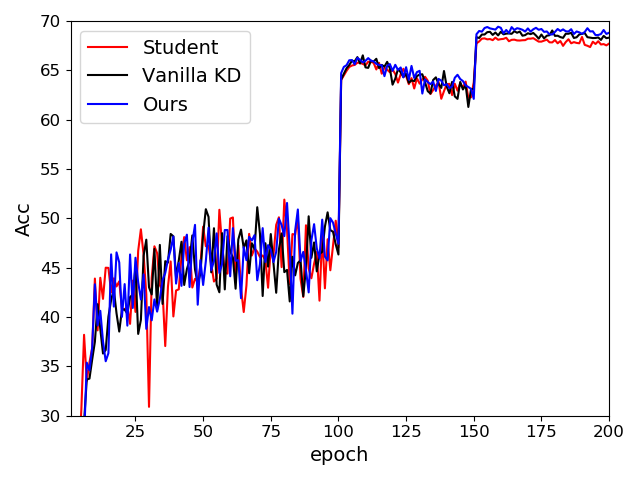}
  }
  \quad \quad
  \subfigure[The student MiCT-ResNet18 on UCF101.]{
  \includegraphics[width=0.4\textwidth]{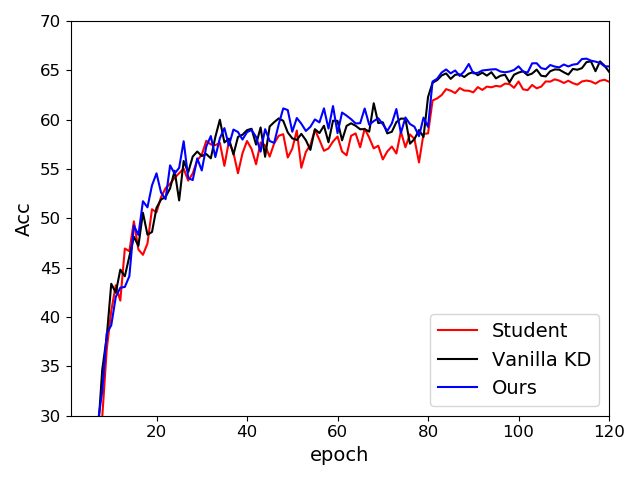}
  }
  
  \caption{Performance curves corresponding to three different learning approaches. The jumps in the curves are caused by the decay of learning rates in SGD~\cite{yuan2019revisit}, which is normal and reasonable.}  \label{fig:curve_Acc}
  \Description{Accuracy of ResNet20 on CIFAR100 and MiCT-ResNet18 on UCF101.}
\end{figure*}

\subsubsection{Performance on Text Modality}
% Implementation Details
Our framework can also be naturally applied to text classification task.
For the two experimental textual datasets, we set the hyperparameters $\alpha=0.5, \beta=0.1$, and temperature $\tau=6.0$.
LW-KD uses $\mathcal{T}_1$ and $\mathcal{T}_2$ as the lightweight teachers for THUCNews and ToutiaoNews, respectively.
By contrast, vanilla KD employs BERT~\cite{devlin2019bert} as the teacher to teach the student models.
We use Adam to update all the student models, including TextCNN, TextRCNN, DPCNN, FastText, and BERT.

Table~\ref{tab:Text-Result} reports the text classification results of the students using different learning methods.
We witness similar phenomena to image classification that distillation based learning methods boost the performance of students, showing the benefits of incorporating distillation learning beyond supervised learning.
And Tf-KD still could not reach the performance level of KD.
More importantly, LW-KD even slightly outperforms KD in most cases of text classification.

\begin{table}[!t]
  \caption{Results of human action classification on UCF101.}
  \label{tab:UCF101-result}
  \resizebox{.95\columnwidth}{!}{\begin{tabular}{c|ccc}
    \toprule
    Method & MiCT-ResNet18& 3D-ResNet18 & 3D-ResNet101 \\\hline
    \#Params & $\sim$16.1M & $\sim$33.3M & $\sim$85.3M \\\midrule
    Sup-Stu & 64.42 & 60.67 & 67.84 \\
    LSR & 65.94 & 61.43 & 67.92 \\
    Tf-KD & 64.98 & 61.12 & 68.01 \\
    KD & 65.97 & 61.91 & -- \\
    Ours[$\mathcal{T}_4$] & \textbf{66.18} & \textbf{62.09} & \textbf{69.17} \\
    \bottomrule
  \end{tabular} }
\end{table}

\subsubsection{Performance on Video Modality}
Finally, we conduct experiments on UCF101 to investigate how well LW-KD works on video modality.
$\mathcal{T}_4$ is leveraged in our learning framework.
We select MiCT-ResNet18~\cite{zhou2018mict}, 3D-ResNet18~\cite{hara2018can}, and 3D-ResNet101~\cite{hara2018can} as students.
For the implementation of KD, we regard 3D-ResNet101 as the teacher of the other two student models.
Considering that all the three student models behave not so well on this task (cf. Sup-Stu in Table~\ref{tab:UCF101-result}), we only train $\mathcal{T}_4$ for 2 epochs to bridge the performance gap between the teacher and the students.
The hyperparameters $\alpha$, $\beta$, and $\tau$ are set to 0.1, 0.05, and 20.0, respectively.
We use SGD for optimization, with an initial learning rate of 0.01 and a decay factor of 0.1. 
Table~\ref{tab:UCF101-result} presents the result details and similar conclusions about LW-KD could be drawn as well.

In a nutshell, the above analysis demonstrates the benefits of LW-KD, for it achieves better classification performance than teacher-free KD and LSR, and could learn from a more efficient lightweight teacher than vanilla KD.
As a complementary, we depict the performance curves w.r.t. specific students on the test sets of CIFAR100 and UCF101.
Figure~\ref{fig:curve_Acc} plots the variation trends, which  could validate the consistent and robust improvement of student models brought by LW-KD, and show the slightly better performance over vanilla KD.

\subsection{Ablation Study of LW-KD (\textbf{\texttt{RQ2}})}

\begin{table}[!t]
  \caption{Ablation study of LW-KD.}
  \label{tab:ablation}
  \resizebox{1\columnwidth}{!}{\begin{tabular}{cccccc}
    \toprule
    Student & Dataset & Ours & w/o ADV & w/o KL & w/o SYN \\
    \midrule
    ResNet20 & CIFAR10 & 92.94 & 92.76 & 92.60 & 92.46 \\
    ResNet20 & CIFAR100 & 69.39 & 69.21 & 69.15 & 68.73 \\
    TextRCNN & THUCNews & 91.86 & 91.48 & 91.53 & -- \\
    TextCNN & ToutiaoNews & 86.65 & 86.28 & 86.42 & -- \\
    MiCT-ResNet18 & UCF101 & 75.62 & 75.22 & 75.05 & -- \\
    \bottomrule
  \end{tabular}  }
\end{table}

This part further analyzes how the main design principles in LW-KD could ensure its effectiveness.
To achieve this, we come up with several variant learning methods of LW-KD.
``w/o ADV'' means removing the adversarial loss from Equation~\ref{eq:Loss}.
``w/o KL'' denotes erasing the KL-divergence loss shown in Equation~\ref{eq:kd-loss}.
``w/o SYN'' represents to train the teacher model LeNetW on the CIFAR datasets, instead of using the synthetic dataset SynMNIST to train the lightweight teacher, just as KD[$\mathcal{T}_5]$ in Table~\ref{tab:s_c10}.

Table~\ref{tab:ablation} describes the corresponding results.
Firstly, we compare ``w/o ADV'' and ``w/o KL'' with the full LW-KD approach and observe that both the two variants suffer from consistent performance drops.
This phenomenon verifies the necessity of incorporating corpus-level knowledge guidance by adversarial loss and instance-level knowledge guidance through KL-divergence loss.
By further investigating ``w/o SYN'', we find that its performance drop is larger than those of the first two variants.
It might be attributed to the fact that the teacher network is a relatively simple model for the CIFAR datasets.
As seen in Table~\ref{tab:letnet5-teacher}, the performance of LeNetW is not very satisfied, compared to results of other more powerful deep networks in Table~\ref{tab:s_c10}.
This confirms the intuition that learning a lightweight teacher on a simple dataset is more reasonable for transferring knowledge to a student as label smoothing.

% -------------------------------------
%% FLOPs
% -------------------------------------
\begin{table*}[!t]
  \caption{Space complexity and computational cost of models on different datasets.}
  \label{tab:efficiency}
  \begin{tabular}{ccc|ccc|ccc}
    \toprule
    \multicolumn{3}{c|}{CIFAR10} & \multicolumn{3}{|c|}{UCF101} & \multicolumn{3}{|c}{ToutiaoNews} \\
    \hline
    Model & \#Params & \#FLOPs & Model & \#Params & \#FLOPs & Model & \#Params & \#FLOPs \\
    \midrule
    *LeNet5 & $\sim$61.7K & $\sim$481.7K & *LeNet5 & $\sim$61.7K & $\sim$481.7K & *LeNet5 & $\sim$69.4K & $\sim$497.0K \\
    *Discriminator & $\sim$4.9K & $\sim$9.7K & *Discriminator & $\sim$4.9K & $\sim$9.7K & *Discriminator & $\sim$10.8K & $\sim$21.4K \\
    ResNet18 & $\sim$11.2M & $\sim$37.2M & MiCT-ResNet18 & $\sim$16.1M & $\sim$12.6G & TextCNN & $\sim$2.1M & $\sim$20.6M \\
    DenseNet121 & $\sim$7.0M & $\sim$898.1M & 3D-ResNet18 & $\sim$33.3M & $\sim$16.8G & FastText & $\sim$152.0M & $\sim$465.9K \\
    ResNeXt29 & $\sim$89.6M & $\sim$14.1G & 3D-ResNet101 & $\sim$85.3M & $\sim$37.7G & BERT & $\sim$102.3M & $\sim$5.4G \\
    \bottomrule
  \end{tabular}  
\end{table*}

\subsection{Efficiency Analysis (\textbf{\texttt{RQ2}})}

To validate the efficiency of LW-KD in training a lightweight teacher, we analyze both the space complexity and computational cost of the relevant models.
In Table~\ref{tab:efficiency}, the first part of results w.r.t. each dataset clarifies the total number of parameters contained in the lightweight teacher LeNet5 and some other adopted teacher networks adopted by KD.
Through systematic comparison, we summarize that the lightweight teacher is about two orders of magnitude smaller than the complicated teachers.
As such, the proposed LW-KD framework is space-efficient in the teacher training stage, which alleviates the issue of large space occupation encountered by vanilla KD.

Moreover, Table~\ref{tab:efficiency} also presents the computational cost in terms of FLOPs needed to generate prediction for each data instance.
This is a reliable measure, no matter how the computational environment changes.
As we can see, the lightweight teacher is about two orders of magnitude faster than the other chosen teachers.
Besides, we also report the cost of the discriminator used in adversarial training in the table.
Compared to the other teachers including the lightweight teacher, the computational cost is nearly negligible.

\section{Conclusions}
In this paper, we study knowledge distillation for neural networks from the perspective of learning a lightweight teacher as label smoothing.
This is motivated by the issues of complicated teachers used in many existing KD approaches and the inflexibility of manually-crafted virtual teachers utilized in teacher-free KD.
We propose the novel distillation learning framework named LW-KD.
In the framework, the lightweight teacher (i.e., LeNet5) is efficiently trained on the simple synthetic dataset SynMNIST, which later generates soft target distribution as teacher knowledge.
The enhanced KD loss is devised to incorporate the knowledge and further guide the effective student learning on a target dataset.
To ensure reliability, we conduct experiments on multiple data modalities.
The comprehensive comparison demonstrates the benefits of LW-KD and validates the rationality of its innovative design principles.

%\begin{acks}
%To Robert, for the bagels and explaining CMYK and color spaces.
%\end{acks}

\bibliographystyle{ACM-Reference-Format}
\bibliography{ref}

%\appendix

% 
% If your work has an appendix, this is the place to put it.
%\clearpage

\appendix

\section{APPENDIX}

To support the reproducibility of our experiments, we detail the architectures of LeNet5 and LeNetW, and the configurations of the compared learning approaches . 

\subsection{Detailed Settings of Model Architectures}

The detailed architecture of the lightweight teacher in our experiments is shown in Table~\ref{tab:arch_lenet5}.
Its original form~\cite{lecun1998gradient} only handles one image channel.
LeNetW is built based on LetNet5 for handling 3 channels, suitable for the RGB images in CIFAR.
Table~\ref{tab:arch_lenetw} shows its concrete architecture, having only 4 layers but being very wide. 

\begin{table}[!h]
  \caption{The architecture of LeNet5 for SynMNIST}
  \label{tab:arch_lenet5}
  \begin{tabular}{c|c}
  \toprule
  \textbf{Output Size} & \textbf{LeNet5} \\
  \midrule
  \multirow{2}*{14 $\times$ 14} & 5 $\times$ 5  6 Conv,ReLU \\
  ~ & maxpool$\downarrow_{2 \times}$ \\
  \hline
  \multirow{2}*{5 $\times$ 5} & 5 $\times$ 5  16 Conv,ReLU \\
  ~ & maxpool $\downarrow_{2 \times}$ \\
  \hline
  1 $\times$ 1 & 5 $\times$ 5  120 Conv,ReLU \\
  \hline
  1 $\times$ 1 & 84 FC \\
  \hline
  1 $\times$ 1 & Output FC \\
  \bottomrule
  \end{tabular}
\end{table}

% Arch of LeNetW
\begin{table}[!h]
  \caption{The architecture of LeNetW for CIFAR}
  \label{tab:arch_lenetw}
  \begin{tabular}{c|c}
  \toprule
  \textbf{Output Size} & \textbf{LeNetW} \\
  \midrule
  30 $\times$ 30 & 3 $\times$ 3 64 Conv,ReLU \\
  \hline
  \multirow{2}*{14 $\times$ 14} & 3 $\times$ 3 128 Conv,ReLU \\
  ~ & maxpool $\downarrow_{2 \times}$ \\
  \hline
  1 $\times$ 1 & 1024 FC, Dropout \\
  \hline
  1 $\times$ 1 & Output FC \\
  \bottomrule
  \end{tabular}
\end{table}

\subsection{Configurations of Approaches}

Table~\ref{tab:config} summarizes the model configurations for each dataset. 
As shown in the second line of the results, we denote `*ALL' as all the models trained on CIFAR10 and CIFAR100, including ResNet20~\cite{he2016deep} and MobileNetV2~\cite{sandler2018mobilenetv2}, etc.

\begin{table*}[h]
  \caption{Detailed configurations of approaches for different datasets.}
  \label{tab:config}
  \begin{tabular}{c|c|c|c|c|c|c|c}
  \toprule
  % Dataset & Model & Input Size & Batch & Optimizer & lr & lr\_step & Epcohs \\
  \textbf{Dataset} & \textbf{Model} & \textbf{Input Size} & \textbf{\#Batch Size} & \textbf{Optimizer} & \textbf{lr} & \textbf{lr\_step} & \textbf{\#Epochs} \\
  \midrule
  SynMNIST, MNIST & LeNet5 & 28 $\times$ 28 & 128 & SGD & 0.1 & -- & 10 \\
  CIFAR & *ALL & 32 $\times$ 32 & 128 & SGD & 0.1 & [100, 150] & 200 \\
  THUCNews, ToutiaoNews & DPCNN, TextCNN, FastText & 1 $\times$ 300 & 128 & Adam & 1e-3 & -- & 20 \\
  THUCNews, ToutiaoNews & TextRCNN & 1 $\times$ 300 & 128 & Adam & 0.1 & -- & 10 \\
  THUCNews, ToutiaoNews & BERT & 1 $\times$ 300 & 128 & Adam & 5e-5 & -- & 3 \\
  UCF101 & MiCT-ResNet18 & 224 $\times$ 224 & 64 & SGD & 0.01 & [80] & 120 \\
  UCF101 & 3D-ResNet18, 3D-ResNet101 & 224 $\times$ 224 & 64 & SGD & 0.01 & [40, 80] & 90 \\
  \bottomrule
  \end{tabular}
\end{table*}

\end{document}